\documentclass[runningheads]{llncs}
\usepackage[T1]{fontenc}
\usepackage{graphicx}
\usepackage{amsfonts, amssymb, amsmath} 
\usepackage{newtxtext,newtxmath} 
\usepackage{subcaption} 

\usepackage{multirow} 

\renewcommand{\problemname}{Mixed-Motive Limited-Observability Legible Motion Planning}
\newcommand{\problemabbrev}{MMLO-LMP}
\newcommand{\algorithmnamelong}{\textbf{DU}al \textbf{BI}ased limited \textbf{O}bservability \textbf{U}nified trajectory \textbf{S}olver}
\newcommand{\algorithmname}{DUBIOUS}

\newcommand{\ie}{\textit{i}.\textit{e}.}

\begin{document}
\title{From Legible to Inscrutable Trajectories: (Il)legible Motion Planning Accounting for Multiple Observers}
\titlerunning{(Il)legible Motion Planning for Multiple Observers}
%
\author{Ananya Yammanuru$^*$ \and
Maria Lusardi$^*$ \and Nancy M. Amato \and Katherine Driggs-Campbell}
\authorrunning{A. Yammanuru, M. Lusardi, et al.}
%
\institute{$^*$ Equal Contribution; Authors listed randomly with seed `$2026$' \\University of Illinois at Urbana-Champaign, Urbana IL 61801, USA}
\maketitle              
\begin{abstract} 
In cooperative environments, such as in factories or assistive scenarios, it is important for a robot to communicate its intentions to observers, who could be either other humans or robots.
A legible trajectory allows an observer to quickly and accurately predict an agent's intention.
In adversarial environments, such as in military operations or games, it is important for a robot to not communicate its intentions to observers.
An illegible trajectory leads an observer to incorrectly predict the agent's intention or delays when an observer is able to make a correct prediction about the agent's intention.
However, in some environments there are multiple observers, each of whom may be able to see only part of the environment, and each of whom may have different motives.
In this work, we introduce the \textbf{\problemname}~(\problemabbrev) problem, which requires a motion planner to generate a trajectory that is legible to observers with positive motives and illegible to observers with negative motives while also considering the visibility limitations of each observer.  
We highlight multiple strategies an agent can take while still achieving the problem objective.
We also present \algorithmname, a trajectory optimizer that solves \problemabbrev.
Our results show that \algorithmname~can generate trajectories that balance legibility with the motives and limited visibility regions of the observers. 
Future work includes many variations of \problemabbrev, including moving observers and observer teaming.  


\end{abstract}
\section{Introduction} 

Modern robotic systems, including warehouse fleets, autonomous vehicles, and patrol agents, often operate alongside other decision-making agents. 
Generally, individual agents are expected to account for the future trajectory of other nearby agents when planning their paths. 
Thus, when an agent moves, it is not only moving closer to its goal, but it is also signaling its intent to observers. 
The task of generating a trajectory that allows an observer to quickly and confidently infer an agent's goal is called \textit{legible motion planning}.  

In this work, we study two special cases of the legible motion planning problem: limited observability and mixed motive. 
Current work in legible motion planning generally optimizes a legibility score over an entire trajectory, thus assuming that an observer can reliably see an entire robot trajectory from start to finish.
In most situations, an observer would not be able to see the agent's full trajectory, either due to obstacles or limited sensor range. 
Though it has been studied in some prior work \cite{taylorObserverAwareLegibilitySocial2022}, further investigation is needed to determine optimal solutions when there are multiple observers simultaneously with different, sometimes overlapping, visibility.

Aside from differences in visibility regions, observers may also have differing motives. 
Most previous work assumes that the observers are cooperative, meaning that the observers' objectives are in-line with the agent's goals, or adversarial, meaning that the observers' objectives are opposed to the agent's goals. 
This implies that it is in the agent's best interest to quickly, or belatedly, reveal its intent while executing a trajectory respectively. 
However, an agent may want to reveal its intention to trusted agents in the scene while simultaneously concealing it from untrustworthy agents. 

These challenges motivate us to introduce the \textbf{\problemname} (\problemabbrev) problem. 
Our goal is to generate trajectories that clearly convey our intent to cooperative agents while concealing it from adversarial agents.
We also present the  \algorithmnamelong~(\algorithmname), which generates trajectories which satisfy the \problemabbrev~requirements. 

While the legible motion planning problem can be solved by optimizing a legibility function, limited observability makes the problem more challenging. 
An observer with a very limited visibility range may see significantly less of a robot trajectory. 
Thus, the agent needs to optimize its time within an observer's visibility region in order to maximize the amount of information conveyed. 
The mixed motive observers add further complexities to this problem. 
A robot's strategy now not only depends on conveying information to its observers, but may also include either avoiding or intentionally misleading a malicious observer. 
If a benevolent agent's and a malicious agent's visibility regions overlap, the robot must appropriately strategize what information conveying. 
This combination of limited observability and mixed motives changes how legibility can be optimized. 
Additionally, the location of a malicious observer's visibility region (near the robot's start position versus near the goals) can change the way an agent should behave.  
In this work, we:
\begin{itemize}
\item Formally define the \problemname~problem 
(\problemabbrev; Section \ref{sec:problemdefinition}). We also present several potential applications for \problemabbrev. (Section \ref{sec:futurework})
\item Present an optimization-based trajectory solver, the \algorithmnamelong~(\algorithmname) for addressing the \problemabbrev~problem. This includes 2 different strategies for addressing a malicious observer. 
(Section \ref{sec:method})
\item Show some examples of interesting \problemabbrev~problem cases, and quantitatively and qualitatively analyze \algorithmname's solutions. 
(Section \ref{sec:results}).
\end{itemize}

\section{Related Work}
There is much preexisting work on legible and illegible planning in both motion and task planning domains.  
Chakraborti et al. \cite{chakrabortiExplicabilityLegibilityPredictability2019} highlighted and differentiated terms used to describe legibility and similar problems, including  "explicability", "predictability", "transparency", or conversely, "security" and "privacy". 
In this section we present work relevant to legible path planning, adversarial path planning, and observers with limited visibility. Then, we briefly discuss how our work addresses all three situations. 

\subsection{Legible Path Planning}
Legible motion planning can be viewed as a subset of implicit communication, where meaning is inferred based on the actions of an agent and context \cite{knepperImplicitCommunicationJoint2017}.
There are many different problem formulations and solutions for legible motion planning depending on what the robot is trying to communicate, the inference model used by the observer, the information available to the observer, and how the trade-off between legibility and efficiency is managed.

In \cite{draganLegibilityPredictabilityRobot2013}, Dragan et al. defined legible motion planning as motion that communicates the robot's true goal from among several possible goals, given that the observer knows the robot's trajectory.
The observer model predicts whichever goal would be most efficiently reached by the observed trajectory.
This model allows legibility to be calculated based on the probability of the true goal given the current trajectory, and optimized via gradient descent \cite{draganGeneratingLegibleMotion2013}.
Legibility is weighted heavier earlier in the trajectory and less heavily later in the trajectory.
Many practical applications of legible motion have been found in social navigation for robots.
Bastarache et al. \cite{Bastarache2023OnLAE} extended the legible motion planning formulation from \cite{draganLegibilityPredictabilityRobot2013} to communicate dynamic goal regions, such as whether a robot intends to pass on the left or right of an oncoming passerby.
Their method allows for real-time path generation by re-planning with motion primitives at each time-step.
Legibility is weighted more heavily in ambiguous situations and efficiency is weighted more heavily otherwise. 
Mavrogiannis et al. \cite{mavrogiannisSocialMomentumFramework2018} integrated a score of "Social Momentum" which takes into account how well a planned action aligns with the predicted avoidance strategies of other agents. 
Geldenbott and Leung \cite{Geldenbott2024LegibleAPA} further considered how the robot's motion will influence the beliefs, and hence the trajectory, of oncoming passerby by using an Iterative Best Response algorithm. 
Che et al. \cite{Che2018EfficientATH} used inverse reinforcement learning to learn a reward model from human behavior to better capture the nuances of human social interaction.

\algorithmname~borrows the formulation of legibility and the observer model from \cite{draganLegibilityPredictabilityRobot2013}. 
This formulation allows us to keep our notion of `goals' abstract, though our method could be extended to dynamic interaction goals as well.
Additionally, we use a cost function for determining legibility instead of a learned reward function to allow greater control in specifying path planning objectives.
However, our work adapts the legibility cost function such that it is not differentiable, making gradient descent methods infeasible.
Instead we use Stochastic Trajectory Optimization for Motion Planning (STOMP) \cite{kalakrishnanSTOMPStochasticTrajectory2011}.
STOMP does not require a differentiable cost function and is less likely to get stuck in local minima.

\subsection{Adversarial Path Planning}
The inverse of legibility is goal "obfuscation" or "privacy" \cite{chakrabortiExplicabilityLegibilityPredictability2019}.
It entails any implicit communication through motion intended to hide some secret about the robot, whether that be about its goals, current location, or other attributes \cite{Kanashima2023FiniteHorizonSFE}. 
Masters and Sardina \cite{mastersDeceptivePathplanning2017} noted that there are two possible strategies for deception: simulation ("showing the false"), and dissimulation ("hiding the true").

Multiple approaches to illegible, or adversarial, path planning have emerged from different disciplines.
Yang et al. \cite{Yang2020SecurebyConstructionOPK} and Kanashima and Ushio \cite{Kanashima2023FiniteHorizonSFE} considered path planning problems defined by Linear Temporal Logic (LTL) constraints. 
Molloy and Nair \cite{Molloy2021SmoothingAverseCCN} considered "covert motion planning" as a type of dynamical system, where the goal is to hide the trajectory from an observer who is using a Bayesian smoother.
Privacy can also be considered in regards to the robot or in regards to other humans in the environment.
Yu et al. \cite{Yu2024PANavTPA} use a Visual Language Model (VLM) to select the most privacy preserving path from among multiple A* generated paths using a top-down view of a space generated by point-cloud data from the robot. 
Paths which are further away from crowded areas or private office spaces are preferred as they limit the robot's exposure to onlookers when completing sensitive tasks, and protect the privacy of other humans from being observed by the robot.

Illegible path planning is also commonly formulated as the direct inverse of legible path planning, \ie~obfuscating the robot's goal given the observer has complete knowledge of the robot's trajectory.
Luo et al. \cite{Luo2019OpponentAwarePWO} implement deceptive motion by planning a path towards a decoy goal up until a Last Obfuscated Turning point, and then plans a path to the intended goal.
Xu et al. \cite{Xu2020ImprovingTSM} formulated deception as a score assignable to each node in a grid based planning problem based on the probability distribution of all possible goals from that node. 
This allows deception to be maximized by reaching the nodes with the highest deceptive values. 
In \cite{draganAnalysisDeceptiveRobot2014}, Dragan et al. presented three methods which extend their original legibility score function to deceptive motion.
Exaggerating motion maximizes legibility to a decoy goal instead of the true goal. 
Switching motion maximizes legibility to a randomly chosen goal, which may include the true goal, at each time step.
Ambiguous motion keeps the probability of all goals equal.
Nichols et al. \cite{Nichols2022AdversarialSMH} developed Adversarial RRT*, which assumes the observer is a trained Recurrent Neural Network (RNN).
The planner uses a separately-trained RNN to generate a probability distribution over all possible goals which are incorporated into the cost as a deception score. 
Deception can be calculated to maximize entropy by rewarding similar probability distributions across all goals, or to maximize simulation by rewarding a high probability of a decoy goal. 
The deception score is added as a component of the cost function used by a sampling-based planner.

We formulate the \problemname~ similar to \cite{draganAnalysisDeceptiveRobot2014,nicholsAdversarialSamplingBasedMotion2022} as it provides a flexible way to plan for legible or illegible scenarios. 
\algorithmname~includes a decoy and an ambiguous strategy analogous to the simulation and dissimulation components identified by \cite{mastersDeceptivePathplanning2017} respectively.
For the decoy strategy, we adapt the model for exaggerating motion as described in \cite{draganAnalysisDeceptiveRobot2014} by selecting a decoy goal to maximize for instead.

\subsection{Limited Visibility Observers}
Limited observability is most often considered as an aspect of illegible path planning algorithms, particularly how it can be exploited.
One solution for illegible path planning is to plan a path that is observed as little as possible.
This is known as the Minimal Exposure Problem \cite{djidjevEfficientComputationMinimum2007}.
There are many variations depending on the type of visibility limitations, such as occlusions from terrain \cite{lebeckComputingHighlyOccluded2013,lebeckComputingHighlyOccluded2014}, or occlusions from limitations of sensor measurement \cite{yuningsongBiologyBasedAlgorithmMinimal2014,fengNovelMinimalExposure2016,djidjevEfficientComputationMinimum2007,veltriMinimalMaximalExposure2003}.
More abstractly, observability can be framed as a graph planning problem, where the goal is to find a path with as few neighboring nodes (\ie~watching observers) as possible \cite{luckowComputationalComplexityLength2020,bevernParameterizedAlgorithmsData2020}, or as polygonal weighted regions in space \cite{buchinGeometricSecludedPaths2020,gewali1988path}.

In legible scenarios, limited observability presents an additional challenge. 
Amirian et al. \cite{Amirian2024LegibotGLB} and Taylor et al. \cite{taylorObserverAwareLegibilitySocial2022} both considered restaurant scenarios, where the observers are customers predicting if a robot is delivering food to their table. 
Visibility is calculated based on the position and orientation of each person, and used to weight a cost function which encourages paths that remain in the field of view.
However, \cite{taylorObserverAwareLegibilitySocial2022} noted the challenge of remaining legible to multiple observers with different overlapping visibility regions.
Dadvar et al. \cite{Dadvar2021JointCAG} considered humans as "noisy sensors" who may not see or misunderstand communication signals from the robot.
They used a search tree to select the optimal combination of explicit communication and path planning while also using Control Barrier Functions to bound collision probability.
Kulkarni et al. \cite{kulkarniUnifiedFrameworkPlanning2019} presented a unified framework for offline task planning which can produce legible or obfuscated plans for observers with partial visibility. 
Observers attempt to learn the agent's true goal from among a set of candidate goals, but only observe observations emitted by the agent's actions instead of the actions directly. 
Some actions emit the same observation, which limits the observer's visibility as to which action was performed.
Agents use this ambiguity to order tasks such that observations are consistent with at least $k$ goals (in an adversarial context) or at most $j$ goals (in a cooperative context).
\cite{kulkarniPlanningProactiveAssistance2021} also considered the problem in the context of proactive assistance.

Like \cite{kulkarniUnifiedFrameworkPlanning2019}, we present a unified method for legible or obfuscated planning, but in a motion planning context.
We treat visibility limitations as limited field of view similar to \cite{Amirian2024LegibotGLB,taylorObserverAwareLegibilitySocial2022}, and model them as polygonal, weighted regions, similar to \cite{buchinGeometricSecludedPaths2020,gewali1988path}.
Similar to \cite{taylorObserverAwareLegibilitySocial2022}, we consider multiple observers at once and how their visibility regions may overlap, but we also consider observers of mixed motive with overlapping visibility regions, adding the challenge of being legible to some observers and illegible to others.
Spending more time in visibility regions of beneficial observers and less time in regions of adversarial observers \cite{Amirian2024LegibotGLB,buchinGeometricSecludedPaths2020,gewali1988path,taylorObserverAwareLegibilitySocial2022} is a strategy which naturally emerged from our cost function. 
However, our method can also generate paths which purposely stay in view of adversarial observers specifically to obfuscate intent further, which in some cases is more effective than staying out of view.

\section{Methodology}
\subsection{Preliminaries and Problem Definition}

We define an environment $E \subset \mathbb{R}^2$ which contains an agent $\mathcal{A}$ and a set of candidate goals $\mathcal{G}$. 
Let $\Xi$ denote the set of all possible trajectories.  
A trajectory $\xi:[0, T]\rightarrow E$ has start position $\xi(0)=S$ and goal position $\xi(T)=G_R$, where $G_R \in \mathcal{G}$. 
Thus, $\xi(t)$ is the agent's position at time $t$. 

\begin{figure*}[t]
\centering
\includegraphics[width=.8\textwidth]{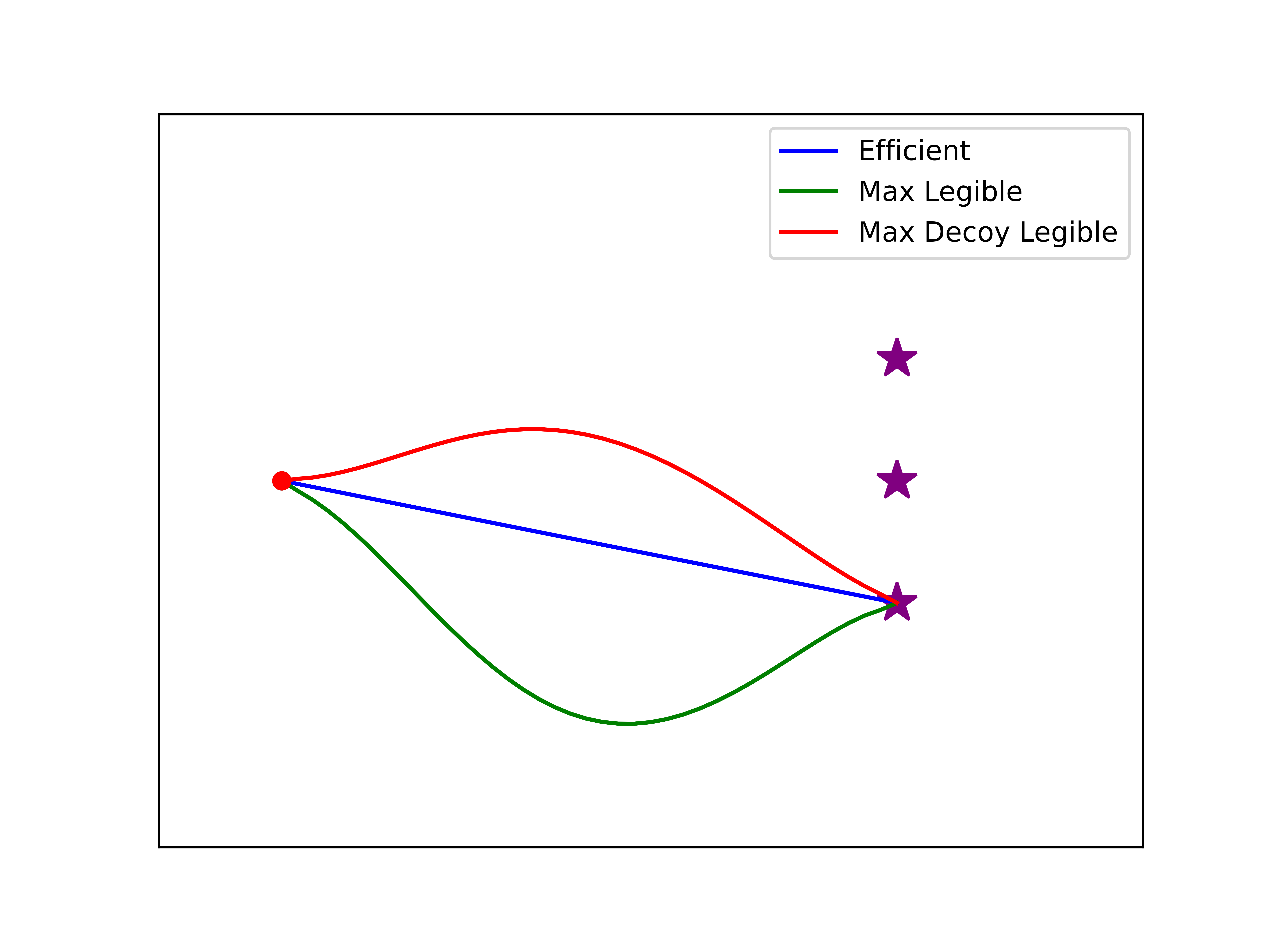}
\caption{An example of a trajectory maximizing legibility to the true goal and a trajectory maximizing legibility to a decoy goal. These baselines are generated with \algorithmname~in an environment that assumes full observability for a single +1 for the max legible trajectory or -1 observer for the max decoy legible trajectory. The decoy trajectory is an example of the $\textsc{Illegible-decoy}$ strategy. Note that the max legible trajectory curves away from the other goal options, reducing the probability that the other goals are the true goal. Conversely, note that the max decoy legible trajectory moves toward the top decoy goal before curving toward its correct goal. }
\label{fig:legible-and-illegible}
\end{figure*}

\subsubsection{Legibility.}

For a trajectory $\xi$, we define the cost $C:\Xi \rightarrow \mathbb{R}^+$ as a measure of efficiency. A lower $C$ indicates a more efficient, or `better' trajectory. For clarity, we refer to $\xi(t)$ for some arbitrary $t$ as $Q$. 

In this work, we use $C(\xi) = \frac{1}{2}\int||\dot{\xi}||^2 dt$ (same as \cite{draganGeneratingLegibleMotion2013}). 
The trajectory which minimizes this cost is a straight line path from $S$ to $G_R$ with constant velocity. 
We also define $C^*_G(Q) = \min_{\xi \in \Xi_{Q\rightarrow G}} C(\xi)$ as the cost of the optimal trajectory from $Q$ to $G$. 

We define \textit{legibility} similarly to \cite{draganAnalysisDeceptiveRobot2014,draganGeneratingLegibleMotion2013,draganLegibilityPredictabilityRobot2013}.
A legible trajectory enables an observer to quickly infer the correct goal based on observing a portion $\xi_{S\rightarrow \xi(t)}$ of a trajectory. 
Thus, we can assign the legibility score of a trajectory as 
\begin{equation} \label{eqn:legscore}
    \textsc{Legibility}(\xi) = \frac{\int P(G_R | \xi_{S \rightarrow \xi(t)}) (T-t) dt}{\int (T-t) dt}
\end{equation}

A high legibility score indicates that $P(G_R | \xi_{S\rightarrow\xi(t)})$ remains high from an early portion of the trajectory. 

It is shown in \cite{draganLegibilityPredictabilityRobot2013} that $P(\xi_{S\rightarrow Q} | G) \approx \frac{k\exp(-C(\xi_{S\rightarrow Q}) - C^*_G(Q))}{\exp(-C^*_G(S))}$ for constant $k$. 
We calculate $P(G_R|\xi_{S\rightarrow Q})$ using Bayes rule and substitution:
\begin{equation}
    P(G_R|\xi_{S\rightarrow Q}) 
= \frac{P(\xi_{S\rightarrow Q}|G_R)P(G_R)}{\sum_{G \in \mathcal{G}}P(\xi_{S\rightarrow Q}|G) P(G)} \label{eqn:1} \\
\end{equation}

There are two ways for a trajectory to increase $P(G_R|\xi_{S\rightarrow Q})$ and hence be more legible. 
First, by being predictable in regards to the true goal (thus increasing $P(\xi_{S\rightarrow Q}|G_R))$. 
Second, by being unpredictable in regards to all other goals (thus decreasing $P(\xi_{S\rightarrow Q}|G)$ for all $G \neq G_R$). 
This can be understood when viewed through the lens of implicit communication \cite{knepperImplicitCommunicationJoint2017}. 
In order for a trajectory to successfully communicate the intended goal, the action must be unpredictable in general, thus triggering the observer to seek an explanation, but more predictable once the true goal is known, thus providing a plausible explanation for the agent's behavior.

\begin{equation}
P(G_R|\xi_{S\rightarrow Q}) 
 \propto \frac{\exp(-C(\xi_{S \rightarrow Q}) - C^*_{G_R}(Q) + C^*_{G_R}(S)) P(G_R)}{\sum_{G \in \mathcal{G}} \exp(-C(\xi_{S \rightarrow Q}) - C^*_{G}(Q) + C^*_G(S) ) P(G)}\label{eqn:probabilityfn}
\end{equation}
An example of a legible trajectory is shown in Fig. \ref{fig:legible-and-illegible}.

\subsubsection{Illegibility.} 
Since legibility is defined as quickly \textit{and} confidently guessing the correct goal after seeing a portion of a trajectory, we can define illegibility as the opposite: either delaying a confidently correct guess for as long as possible, \textit{or} quickly and confidently guessing the wrong goal.

We penalize early correctness and encourage incorrectness  with the following equation:

\begin{equation} 
\label{eqn:illegscore}
    \textsc{Illegibility}(\xi) = \frac{\int (1-P(G_R | \xi_{S \rightarrow \xi(t)})) (T-t) dt}{\int (T-t) dt}
\end{equation}

Coincidentally, note:
\begin{equation}
\label{eqn:addto1}
\textsc{Legibility}(\xi) + \textsc{Illegibility}(\xi) = 1
\end{equation}

An example of an illegible trajectory (using the decoy strategy) is shown in Fig. \ref{fig:legible-and-illegible}.

\subsection{\problemname.}
\label{sec:problemdefinition}
Now, consider that our environment also has observers $\mathcal{O}$.
Each observer $o \in \mathcal{O}$ has a visibility region $V_o\subseteq E$ and motive $M_o\in[-1, 1]$.
For simplicity, we refer to observers with $M_o < 0$ as "negative" observers and those with $M_o \geq 0$ as "positive" observers. 

We define $\nu_{o}:[0, T] \rightarrow \Xi$, where $\nu_o(i)$ is the portion of $\xi_{S\rightarrow \xi(i)}$ that lies within $V_o$. Also, let $T_o$ be the number of timesteps during which $\xi$ lies in $V_o$. 

The goal of the \problemname~problem (\problemabbrev) is to find a  trajectory $\xi$ from $S$ to $G_R$ such that $\xi$ is legible to positive observers but is illegible to negative observers. In mathematical terms, we want to find $\xi$ such that:
\begin{equation}\label{eqn:mmlolmp-trajectory-definition} \xi = \arg\max_{\xi \in \Xi} \sum_{o \in \mathcal{O}, M_o \geq 0} M_o \times \textsc{Legibility}(\nu_{o}(T)) + \sum_{o \in \mathcal{O}, M_o < 0} |M_o|\times \textsc{Illegibility}(\nu_{o}(T)) 
\end{equation}

If $\xi$ is empty, then we assume $P(G_R|\xi) = 0$. We list a few variants of the \problemabbrev~problem in Section \ref{sec:futurework}. 

\subsection{Method}
\label{sec:method}
We propose a trajectory optimization-based approach called \algorithmnamelong~(\algorithmname) to solve \problemabbrev.
We begin with $\xi_0$ as the straight line trajectory from $S$ to $G_R$. To maximize legibility, we iteratively update $\xi$ using STOMP \cite{kalakrishnanSTOMPStochasticTrajectory2011} with the cost function $\mathcal{F}$ presented below. 
This function is not only designed to maximize legibility or illegibility as defined by the \problemabbrev~statement, but also to balance the trajectory construction in regions where observers with opposing motives overlap.

Since STOMP requires a cost for each point  in the trajectory, $\mathcal{F}$ takes a timestep $i$ as input and uses $\xi_{S\rightarrow \xi(i)} $ to calculate costs. 

We first consider the positive observers. 
We need to maximize legibility for each observer in only what lies within their visibility region. Thus, we re-formulate the legibility function and use the observer's motive as a scaling factor: 

\begin{equation}\label{eqn:fpos}
\mathcal{F}_{pos}(i) = \sum_{o \in \mathcal{O}, M_o \geq 0 } M_o \frac{\int P(G_R|\nu_o(i)) (T_o-t) dt}{\int (T_o-t) dt}
\end{equation}

That is, for each observer, the legibility score only considers points within that observer's visibility region. 

Then, we consider the negative observers. We propose two strategies that an agent can follow: the decoy strategy and the ambiguous strategy.

The decoy strategy misleads the observer to quickly and confidently predict the agent is moving towards a decoy goal instead of the true goal. 
We can measure this strategy similarly to \cite{draganAnalysisDeceptiveRobot2014} for some $G_D \neq G_R$ as:
\begin{equation}\label{eqn:illeg-decoy-score}
    \textsc{Illegibility-decoy}(\xi) = \frac{\int P(G_D|\xi_{S \rightarrow \xi(t)}) (T-t) dt}{\int (T-t) dt}
\end{equation}

The ambiguous strategy maintains ambiguity among multiple candidate goals by minimizing the average difference between $P(G_R|\xi)$ and $P(G|\xi)$ for all other $G \in \mathcal{G} \setminus \{G_R\}$.  
This extends the method in \cite{draganAnalysisDeceptiveRobot2014} to an arbitrary number of goals.
So:
\begin{equation}\label{eqn:illeg-ambig-score}
\textsc{Illegibility-ambiguous}(\xi) = \frac{\int (T-t)(1 - \frac{1}{|\mathcal{G}|}\sum_{G \in \mathcal{G} \setminus \{G_R\}} |P(G_R|\xi) - P(G|\xi)| ) dt}{\int(T-t) dt}
\end{equation}


We construct a cost function for negative observers similar to the one for positive observers, however, using the decoy goal $G_D$:
\begin{equation}\label{eqn:fneg}
\mathcal{F}_{neg}(i) = \sum_{o \in \mathcal{O}, M_o < 0 }  \alpha_{neg}|M_o| \frac{\int P(G_D|\nu_o(i)) (T_o-t ) dt}{\int (T_o-t) dt}
\end{equation}
Even though the illegibility score above uses the better of the decoy and ambiguous strategies, \algorithmname~does not do these calculations. 
Rather, we leave the strategy type as a hyperparameter  $\alpha_{neg}$. 
If $\alpha_{neg} = 1$, then the agent maximizes legibility of the decoy goal for the observer. If $\alpha_{neg} = -1$, then the negative opponent's visibility region is costly to navigate through, and thus the agent avoids the negative opponent. We note that an avoidance strategy is an extremely efficient interpretation of the illegibility ambiguous strategy in limited-observability environments. 
This strategy is also quite similar to the approach of the Minimum Exposure Problem \cite{djidjevEfficientComputationMinimum2007}.

While we have integrated motives into the costs above, they must be normalized over the total number of observers who can see $\xi({i})$. And lastly, STOMP's goal is to minimize cost, so we flip the sign on our cost function. Thus, our final cost function $\mathcal{F}$ is:
\begin{equation}\label{eqn:F-cost}
\mathcal{F}(i) = -\frac{\mathcal{F}_{pos}(i) + \mathcal{F}_{neg}(i)}{\sum_{o\in\mathcal{O}}\mathbb{I}_{q_i\in V_o}}
\end{equation}


We optimize $\xi$ using $\mathcal{F}$ with STOMP, initialized with the most predictable trajectory as defined in \cite{draganGeneratingLegibleMotion2013}. With our cost function, this is the straight-line trajectory from $S$ to $G_R$. For this work, we assume that $P(G)$ is uniformly distributed across all goals.

\begin{figure*}[t]
\centering  
\includegraphics[width=0.75\textwidth]{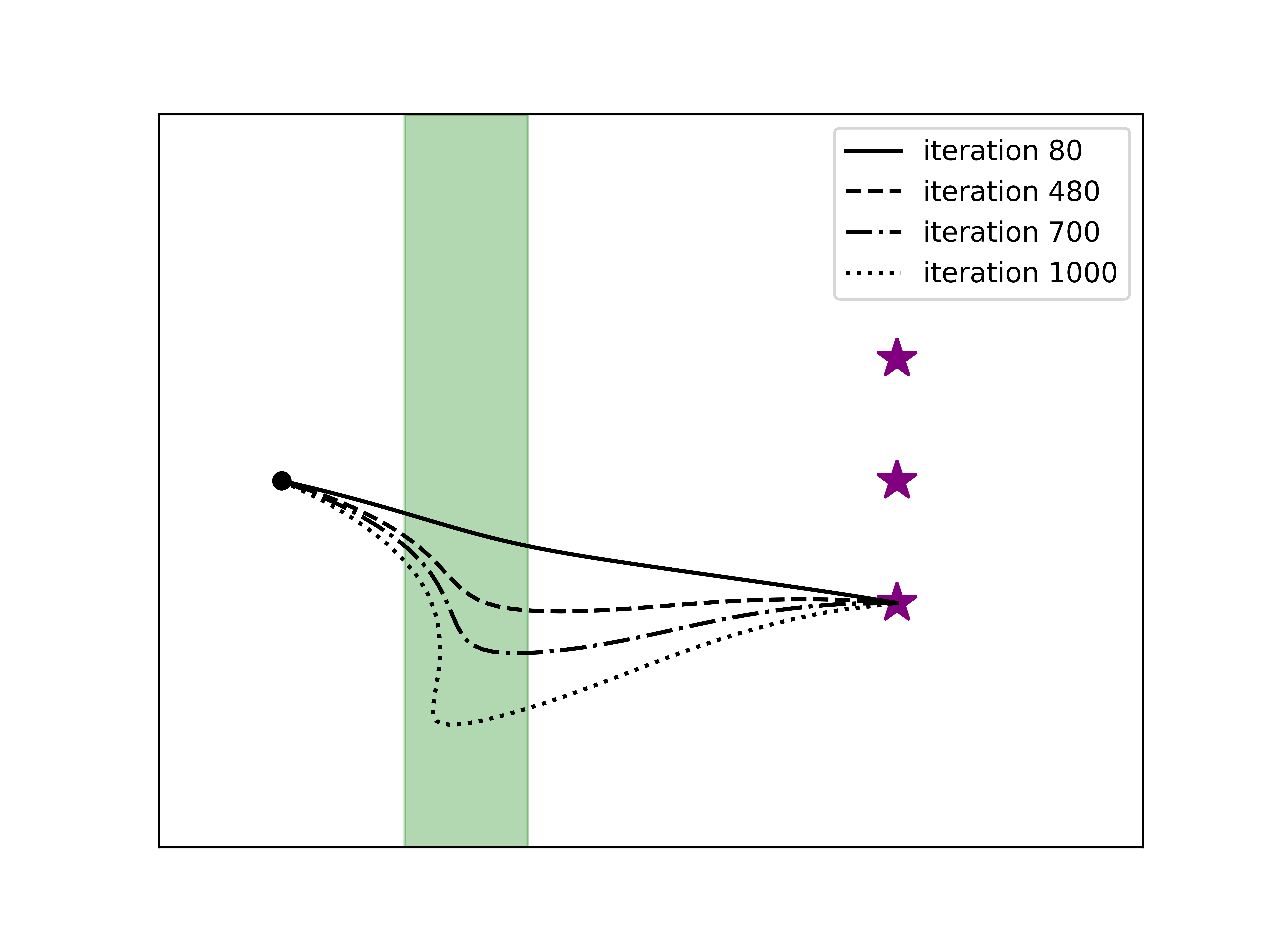}
\caption{STOMP Over-optimization. 
As the number of iterations increases, STOMP creates increasingly extreme trajectories. Here we show legible trajectories generated through the region in view of an observer with motive (green) +1 at iterations 80, 480, 700, and 1000.}
\label{fig:stomp_iterations}
\end{figure*}

\section{Experiments \& Results}
\label{sec:results}

In this section, we present examples of the \problemabbrev~problem, and we demonstrate that \algorithmname~can produce trajectories that achieve the problem objective. 
The environments we use admit multiple strategies. 
We compare \algorithmname~against the full-environment legible and illegible trajectories (from Fig. \ref{fig:legible-and-illegible}). 
We find that \algorithmname~can produce multiple successful strategies and can adapt its trajectories based on observer motive, visibility and goal location. 

Our optimization method, STOMP \cite{kalakrishnanSTOMPStochasticTrajectory2011}, relies on multiple hyperparameters for path generation.
For our experiments we tuned our hyperparameters to produce overall best results. 
Unless otherwise stated, all paths were generated with $40$ way-points, and $1000$ STOMP iterations. STOMP parameters include $K=100$ noisy trajectories, $\lambda=0.1$, and dividing $\mathbf{R}^{-1}$ by $10000$.
STOMP can be run until `convergence'; however, for consistency we run each optimization for $1000$ iterations. 
We acknowledge the need for hyperparameter tuning is a limitation of our method and propose further investigation of optimal trajectory generation methods as future work.


\subsubsection{Evaluation Metrics.} 
For all cases, we qualitatively analyze our trajectories. 
We also report $P(G_i|\xi_{S\rightarrow \xi(t)})$ for each goal, over time, as seen by each observer. These are shown in the plots below each trajectory. 
Additionally, 
we report $\textsc{Legibility}$ score (equation \ref{eqn:legscore}). Because of equation \ref{eqn:addto1}, we do not need to report $\textsc{Illegibility}$. For comparisons in negative observer regions, we report {\scshape Illegibility-ambiguous} and $\textsc{Illegibility-decoy}$ (Equations \ref{eqn:illeg-decoy-score} and \ref{eqn:illeg-ambig-score}, respectively). 
All of these scores are calculated with the probability function from Equation \ref{eqn:probabilityfn}. 

It is important to note that these numbers are intended to be compared against each other. For example, the highest possible score for $\textsc{Legibility}$ is 1; however, this requires $P(G_R|\xi_{S\rightarrow\xi(t)}) = 1$ for every possible value $t$. This is only true when there is 1 possible goal. The same applies for $\textsc{Illegibility-decoy}$. The maximum value of $\textsc{Illegibility-ambiguous}$ is $\frac{1}{|\mathcal{G}|}$, as demonstrated, unless the trajectory is never seen by the observer, in which case, it can be 1 as seen in Table \ref{tab:single-observer}. 

\subsubsection{Legibility Strategy.}


\begin{figure}[t!]
\centering
\begin{subfigure}[t]{0.45\textwidth}
    \center
    \includegraphics[width=\linewidth]{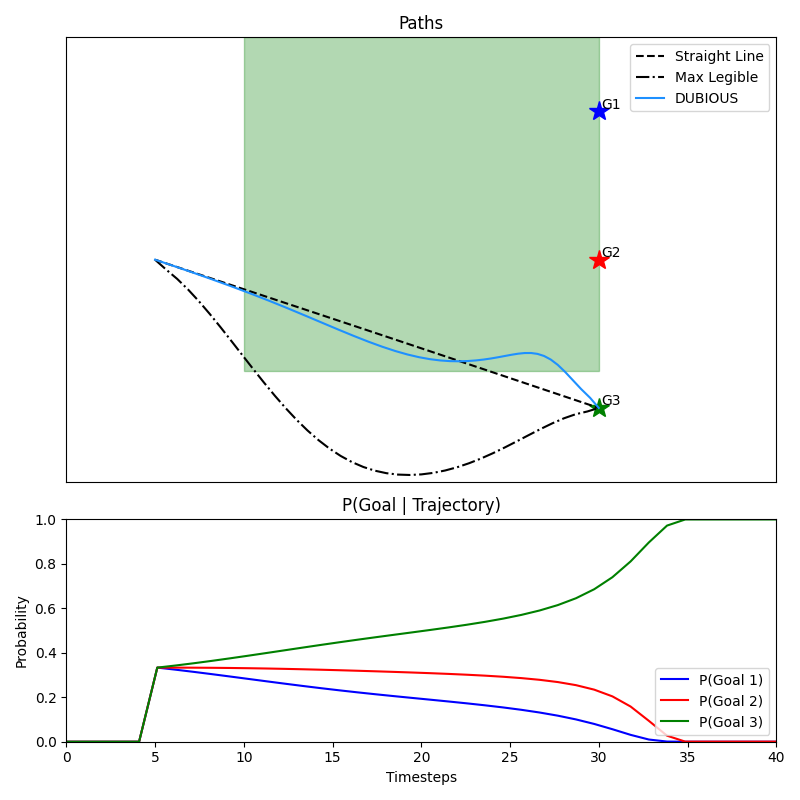}
    \caption{+1 motive observer in green region and observer probabilities}
    \label{fig:partial_view:leg}
\end{subfigure}
\begin{subfigure}[t]{0.45\textwidth}
    \centering
    \includegraphics[width=\linewidth]{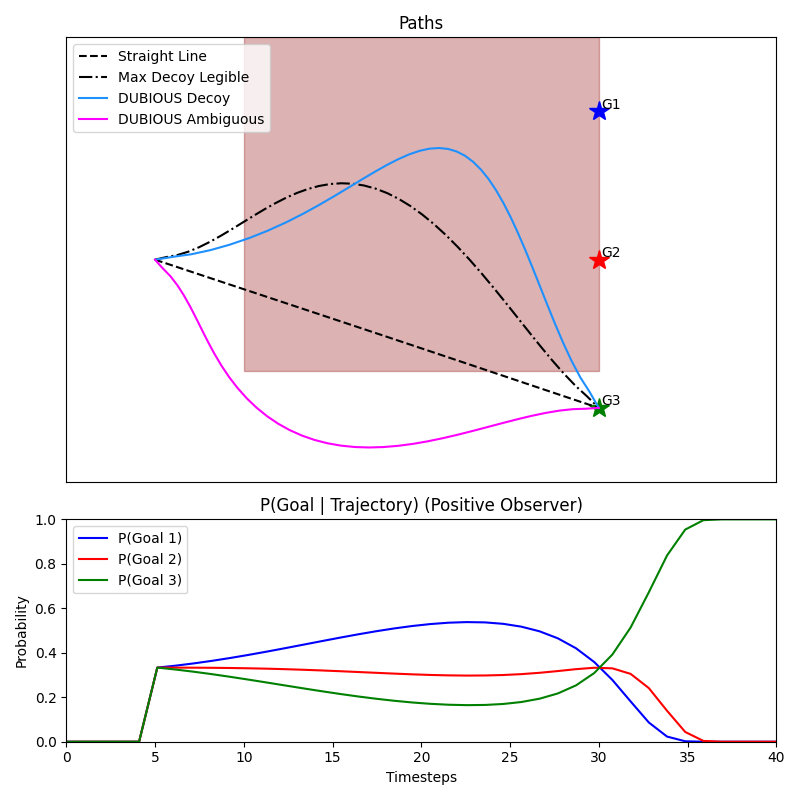}
    \caption{-1 motive observer in red region and probabilities}
    \label{fig:partial_view:illeg}
\end{subfigure}

\caption{(Top) Environments with one observer with partial legibility. 
Fig. \ref{fig:partial_view:leg} has one observer with +1 motive that can see the agent's position when it is in the green box.  
Fig. \ref{fig:partial_view:illeg} has one observer with -1 motive that can see the agent's position when it is in the red box. 
(Bottom) Probabilities $P(G_i|\xi_{S\rightarrow\xi(t)})$ for each goal, over time, as seen by the observer. Fig. \ref{fig:partial_view:leg} shows goal probabilities for the DUBIOUS legible strategy and Fig. \ref{fig:partial_view:illeg} shows goal probabilities for the DUBIOUS decoy strategy. The ambiguous trajectory does not enter the observer's visibility region, so its goal probabilities remain at 0 for all time-steps. 
}
\label{fig:partial_view}
\end{figure}

Fig. \ref{fig:partial_view} shows \algorithmname's performance in environments with a single observer, with either +1 or -1 motive, and partial observability.
Fig. \ref{fig:partial_view:leg} shows a +1 motive observer that can see the green portion of the environment. 
We show 3 candidate trajectories: the "Efficient" trajectory ("Straight Line"), which minimizes $C(\xi)$, the maximally legible trajectory as calculated for Fig. \ref{fig:legible-and-illegible}, and a trajectory generated by \algorithmname. 

The maximally legible trajectory only optimizes for legibility and will not consider the observer, thus inadvertently avoiding the observer. 
As a result, the observer would not have enough information for goal prediction. 
In this case, the efficient straight line trajectory provides the observer with more information since it goes through its visibility region.
\algorithmname, however, employs legible motion while maximizing its time in the observer's view, enabling higher confidence in the correct goal earlier in the trajectory (see Table \ref{tab:single-observer}). 
The bottom portion of Figure \ref{fig:partial_view:leg} shows the probability distribution for $P(G_i|\xi_{S\rightarrow \xi(t)})$ for each goal as time progresses across the trajectory generated by \algorithmname. 
We see that our trajectory distinguishes Goal 3 as the most likely goal early in the trajectory.  
However, given the observer's viewpoint, any path that maximizes time in that region will be more in the direction of the incorrect goals. 
For this reason, despite producing a higher probability of the correct goal, our method has a legibility score only $0.012$ higher than the efficient strategy.
Our method does not determine how much time spent in the observer's view is optimal for legibility or illegibility. 
We leave determining the ideal trade-off for future work.



\begin{table}[h]
\centering
\caption{Metrics for all experiments. Most legible values are highlighted in bold for positive observers. Least legible (most illegible) values are italicized for negative observers. These scores are calculated over the portions of the trajectory that are visible to each observer. }
\label{tab:single-observer}
\begin{tabular}{|l|l|l|l|l|l|}
\hline
                                                                       &                    \multicolumn{2}{c|}{}& $\textsc{Legibility}$ & $\textsc{Illeg-Decoy}$ & $\textsc{Illeg-ambiguous}$ \\ \hline
\multirow{3}{*}{Figure \ref{fig:partial_view:leg}}   & \multicolumn{2}{l|}{Straight Line}& 0.440& ---& ---\\ \cline{2-6}& \multicolumn{2}{l|}{Max Legible}& 0.333& ---& ---\\ \cline{2-6}& \multicolumn{2}{l|}{DUBIOUS}& \textbf{0.452}& ---& ---\\ \hline
\multirow{4}{*}{Figure \ref{fig:partial_view:illeg}} & \multicolumn{2}{l|}{Straight Line}& 0.440& 0.128& 0.864\\ \cline{2-6}& \multicolumn{2}{l|}{Max Decoy Legible}& 0.349& 0.189& 0.933\\ \cline{2-6}& \multicolumn{2}{l|}{DUBIOUS Decoy}& 0.266& \textit{0.313}& ---\\ \cline{2-6}& \multicolumn{2}{l|}{DUBIOUS Ambiguous}& \textit{0}& ---& \textit{1}\\ \hline
\multirow{6}{*}{Figure \ref{fig:overlap_decoy} and \ref{fig:overlap_ambig}}& \multirow{2}{*}{Straight Line}&$+1$ Observer& 0.438& ---& ---\\ \cline{3-6}& &$-1$ Observer& 0.438& \textit{0.057}& 0.958\\
 \cline{2-6}& \multirow{2}{*}{DUBIOUS Decoy}&$+1$ Observer& 0.539& ---&---\\
 \cline{3-6}& &$-1$ Observer& 0.294& 0.021&---\\ 
 \cline{2-6}& \multirow{2}{*}{DUBIOUS Ambig.}&$+1$ Observer& \textbf{0.540}& ---& ---\\ \cline{3-6}& &$-1$ Observer& \textit{0}& ---& \textit{1}\\ \hline
\multirow{2}{*}{Figure \ref{fig:opp-goals}}& \multicolumn{2}{l|}{Straight Line}& 0.676& 0.069& ---\\
 \cline{2-6}& \multicolumn{2}{l|}{DUBIOUS Decoy}& \textbf{0.234}& \textit{0.086}&---\\ \hline
\multirow{4}{*}{Figure \ref{fig:box-in-box:good-in-bad}}& \multirow{2}{*}{Straight Line}&$+1$ Observer& \textbf{0.431}& ---& ---\\ \cline{3-6}& &$- 0.25$ Observer& 0.442& 0.190& ---\\
 \cline{2-6}& \multirow{2}{*}{DUBIOUS Decoy}&$+1$ Observer& 0.390& ---&---\\
 \cline{3-6}& &$- 0.25$ Observer&\textit{ 0.400}& \textit{0.227}&---\\ \hline
\multirow{4}{*}{Figure \ref{fig:box-in-box:bad-in-good}}& \multirow{2}{*}{Straight Line}&$+0.25$ Observer& \textbf{0.442}& ---& ---\\ \cline{3-6}& &$-1$ Observer& 0.431& 0.051& ---\\ 
 \cline{2-6}& \multirow{2}{*}{DUBIOUS Decoy}&$+0.25$ Observer& 0.369& ---&---\\
 \cline{3-6}& &$-1$ Observer& \textit{0.368}& \textit{0.152}&---\\ \hline
\end{tabular}
\end{table}

\subsubsection{Illegible Decoy Strategy}
Fig. \ref{fig:partial_view:illeg} shows 4 trajectories: the efficient trajectory, the maximum decoy legibility as calculated for Fig \ref{fig:legible-and-illegible}, a \algorithmname~trajectory with $\alpha_{neg}=1$ (`Decoy'), and a \algorithmname~trajectory with $\alpha_{neg}=-1$ (`Ambiguous'). 


\algorithmname~can employ two different strategies for handling negative observers.
The decoy strategy initially moves toward a decoy goal, and changes its trajectory as late as possible.
The ambiguous strategy attempts to minimize time spent in the observer's region. Thus, in this example, it moves down to go below the observer's visibility region. 
The bottom of Figure \ref{fig:partial_view:illeg} shows the observer's goal guesses over the course of both of \algorithmname's trajectories. 
The decoy strategy spends longer in the observer's region, but does not provide definitive information to the observer until it begins to turn. 
The ambiguous strategy never enters the observer's region. 
Since the observer never sees the agent, the probabilities across all goals remain 0, so only probabilities for the decoy strategy are plotted.
Table \ref{tab:single-observer} validates the efficacy of both strategies in this environment.

\subsubsection{Illegible Ambiguous Strategy}
\begin{figure}[t!]
\centering
\begin{subfigure}[t]{0.45\textwidth}
    \center
    \includegraphics[width=\textwidth]{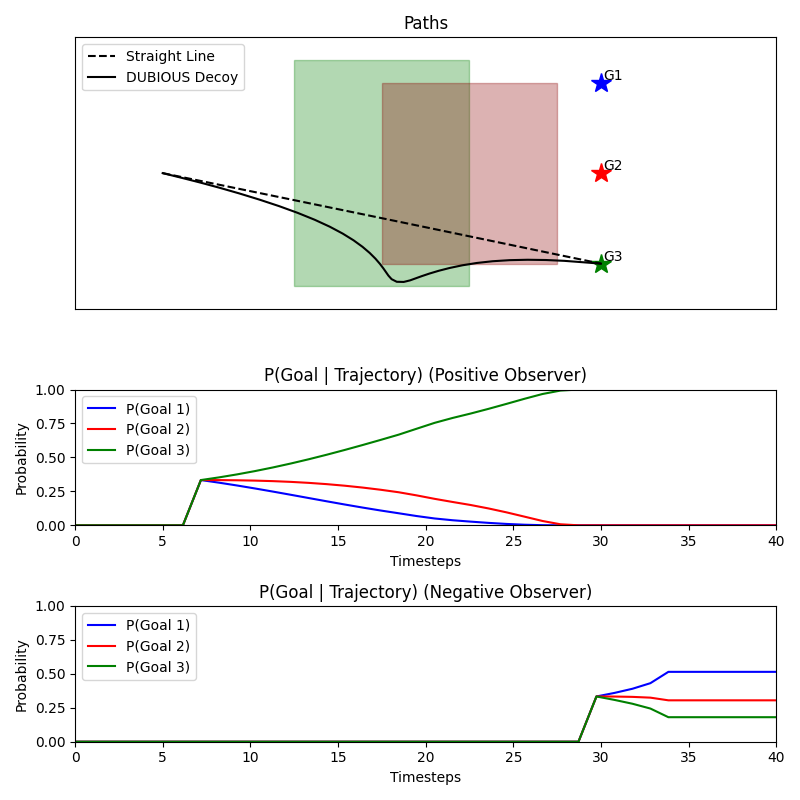}
    \caption{DUBIOUS-decoy path and observer probabilities. }
    \label{fig:overlap_decoy}
\end{subfigure}
\begin{subfigure}[t]{0.45\textwidth}
    \centering
    \includegraphics[width=\linewidth]{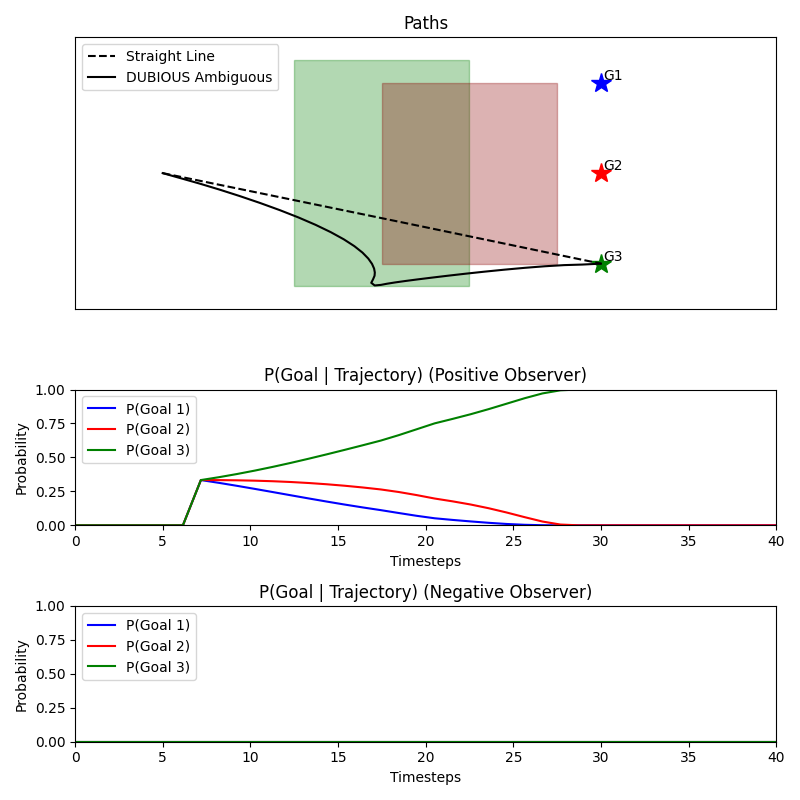}
    \caption{DUBIOUS-ambiguous path and observer probabilities}
    \label{fig:overlap_ambig}
\end{subfigure}

\caption{(Top) Environments with two observers with partial and overlapping legibility. 
(Bottom) Probabilities $P(G_i|\xi_{S\rightarrow\xi(t)})$ for each goal and each observer over time. The ambiguous trajectory does not enter the observer's visibility region, so its probabilities remain 0. 
}
\label{fig:illeg_ambig_strategy}
\end{figure}

There are some scenarios where using an ambiguous or avoidance strategy is more effective than a decoy strategy. 
Figure \ref{fig:illeg_ambig_strategy} shows an environment with two overlapping observers in front of the goals, one positive and one negative. It is possible for the agent to bypass one or both regions in order to reach its goal. 
Figure \ref{fig:overlap_decoy} shows the agent using the decoy strategy traveling through the positive observer's region, providing as much information as possible about its goal. 
Then, it briefly travels through the negative observer's visibility region, giving the negative observer information to infer the true goal is goal 1. 
The agent leaves the visibility region quickly enough that the observer's inference does not change. 
The trajectory in Figure \ref{fig:overlap_ambig} travels through the positive observer's region in a similar manner, but skips the negative observer's region completely. This trajectory provides absolutely no information to the negative observer.  
While the decoy agent has successfully misled the negative observer with its trajectory, the ambiguous agent's strategy is more successful by avoiding the negative region completely. 
Table \ref{tab:single-observer} quantitatively reaffirms that the ambiguous strategy returns the path that is most legible to the positive observer and least legible to the negative observer. 
It is important to note that the straight line trajectory has a higher \textsc{Illegibility-decoy}, because it spends more time in the negative observer's region. 
This experiment also demonstrates that different environment and observer configurations may be more suited to either a decoy or ambiguous strategy. 
Further work includes more detailed analysis on when each strategy is more  successful. 

\subsubsection{Goals in Opposite Directions}
\begin{figure*}[t]
    \centering
    \includegraphics[width=0.6\textwidth]{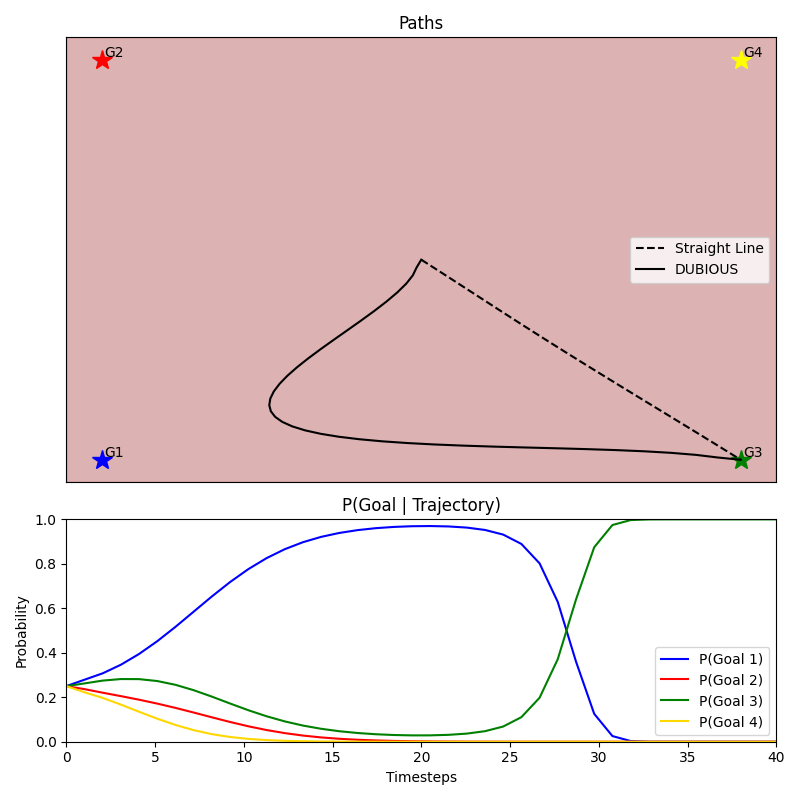}
    \caption{Decoy strategy on environment with potential goals in opposite directions. The agent goes toward decoy goal 1 before pivoting to the true goal later in the trajectory.}
    \label{fig:opp-goals}
\end{figure*}

Figure \ref{fig:opp-goals} illustrates a challenging case where all potential goals are in opposing directions.
The most legible trajectory and the most ambiguous trajectory would both be the straight line shortest path in this scenario.
The straight line path is completely dissimilar from the most efficient path to any other goal, making it legible. 
Since there is no way to avoid the negative observer's visibility region, the ambiguous strategy will try to minimize time spent there, and thus time spent on the path in total.
Therefore it will also choose the straight line path.
The decoy strategy, however, elicits interesting behavior. 
The opposing directions of the goals make it easier to maximize the probability of the decoy goal.
Indeed, for several time-steps at the middle of the trajectory, the observer is near 100\% confident that the decoy is the true goal.
It is not until almost timestep 30 that the observer shifts to predicting the correct goal.
This makes the decoy method ideal for adversarial planning in such scenarios.


\subsubsection{Overlapping Visibility Regions. }


\begin{figure*}[h]
\centering
\begin{subfigure}[t]{0.45\textwidth}
    \centering
    \includegraphics[width=\textwidth]{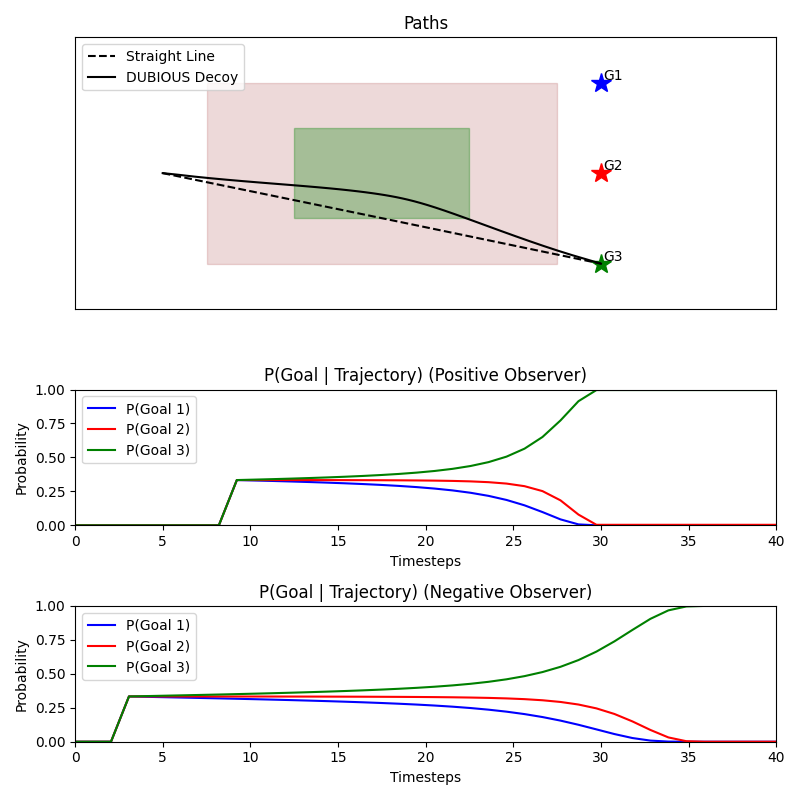}
    \caption{+1 observer within -0.25 observer}
    \label{fig:box-in-box:good-in-bad}
\end{subfigure}
\begin{subfigure}[t]{0.45\textwidth}
    \centering
    \includegraphics[width=\textwidth]{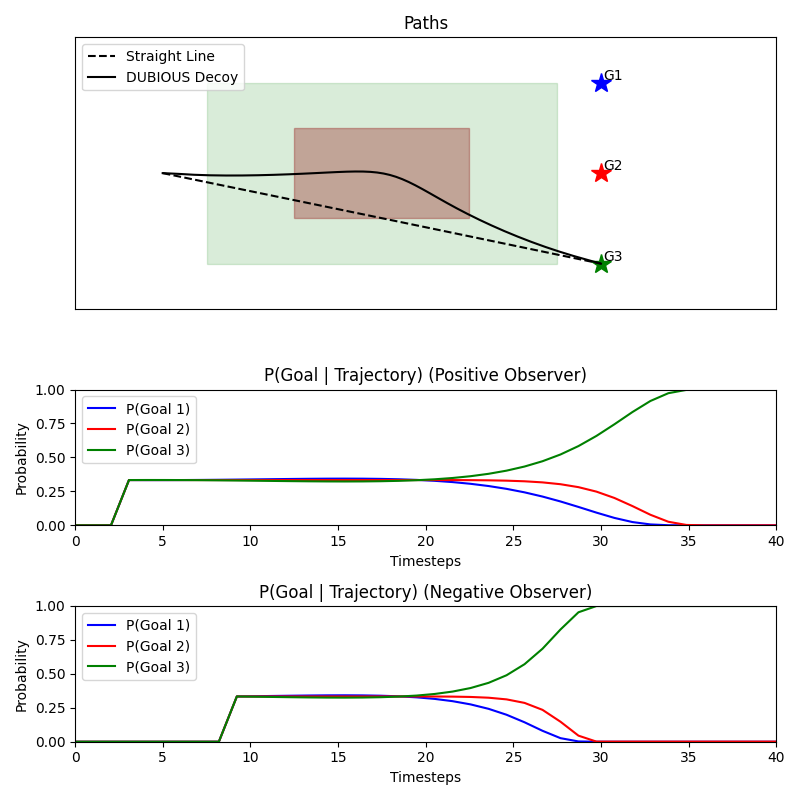}
    \caption{-1 observer within +0.25 observer}
    \label{fig:box-in-box:bad-in-good}
\end{subfigure}
\caption{Both figures illustrate multi-observer environments with observers of different motive values and overlapping visibility regions. 
In Fig. \ref{fig:box-in-box:good-in-bad}, the visibility region of a +1 observer is completely covered by the visibility region of a -0.25 observer. 
In Fig. \ref{fig:box-in-box:bad-in-good}, the visibility region of a -1 observer is completely covered by the visibility region of a +0.25 observer.
}
\label{fig:box-in-box}
\end{figure*}

Figures \ref{fig:illeg_ambig_strategy} and \ref{fig:box-in-box} show environments with two observers with opposing motives and overlapping visibility regions. 
For \algorithmname~trajectories in Figure \ref{fig:box-in-box} we show only the decoy strategy for both experiments.
Fig. \ref{fig:box-in-box:good-in-bad} shows a visibility region of an observer with motive 1 within the visibility region of an observer with motive -0.25. 
Fig. \ref{fig:box-in-box:bad-in-good} shows a visibility region of an observer with motive -1 within the visibility region of an observer with motive 0.25.
In overlapping regions, \algorithmname~weights the legibility score of each observer according to their motive. 
In Fig. \ref{fig:box-in-box:good-in-bad}, the +1 observer dominates the -0.25 observer, and thus the trajectory makes a weak move toward a decoy goal. 
In Fig. \ref{fig:box-in-box:bad-in-good}, the -1 observer's motive dominates the +0.25 observer, resulting in a larger move toward a decoy goal. 
Quantitatively, between these two environments and their baseline, the \textsc{Illegibility-decoy} score significantly increases relative to the strength of the negative observer's motive. However, the introduction of the negative observers with fully overlapping visibility regions resulted in a tradeoff: legibility scores are lower than the baseline in positive observer regions due to the optimizer's attempt to balance the two observers' motives. 

\section{Potential Variants of \problemabbrev}
\label{sec:futurework}
Here, we presented a simplified version of the \problemabbrev~problem. 
Since this is a novel problem, there are many interesting variations. 
In this section, we detail the directions and potential applications of \problemabbrev.  

\subsection{Scenario Modifications}
Our current formulation is a bounded 2-D environment with no obstacles, but in order to bring \algorithmname~solutions closer to the real-world situations, obstacles are a necessary component. 
Aside from collision avoidance challenges, obstacles provide visibility region complexities and new potential strategies, such as hiding behind obstacles to avoid negative observers.
Previous work addressing the Minimum Exposure Problem \cite{djidjevEfficientComputationMinimum2007,lebeckComputingHighlyOccluded2013,lebeckComputingHighlyOccluded2014,yuningsongBiologyBasedAlgorithmMinimal2014,fengNovelMinimalExposure2016,veltriMinimalMaximalExposure2003} could be incorporated into \algorithmname~to improve path planning in these scenarios.
The problem formulation can also include a task planning component. 
An example of a task planning extension to \problemabbrev~includes providing the agent with a list of goals that it has to reach \cite{Yang2020SecurebyConstructionOPK,Kanashima2023FiniteHorizonSFE}, or explicit communication actions it could take \cite{Dadvar2021JointCAG,Che2018EfficientATH}.
The agent would need to plan both the goal ordering and the trajectory such that an observer would need to be able to infer both.

\subsection{Observer Modifications}
We do not consider dynamic observers in our current formulation of \problemabbrev, and we assume visibility regions and motives of each observer are known and static.
However, in many scenarios, such as social navigation, the observers are also agents acting in the environment.
This introduces additional complications, as acting observers may have changing visibility regions and objectives, and neither may be known at the start. 
Many works consider limited observability in terms of environment visibility \cite{lebeckComputingHighlyOccluded2013,lebeckComputingHighlyOccluded2014,Amirian2024LegibotGLB,taylorObserverAwareLegibilitySocial2022} and sensor limitations \cite{yuningsongBiologyBasedAlgorithmMinimal2014,fengNovelMinimalExposure2016,djidjevEfficientComputationMinimum2007,veltriMinimalMaximalExposure2003}, but both factors may jointly impact an observer's total visibility.
For example, the boundaries of an observer's visibility may be uncertain due to sensor uncertainty, resulting in more certain observations in the middle of the region and less certain observations at the boundaries.
The observer may be concealing where their regions of visibility are (similar to hiding linear constraints in constrained policy optimization \cite{benvenutiGuaranteedFeasibilityDifferentially2024}), requiring an agent to infer their locations.
Active observers could take actions to increase their information gain, requiring the agent to adapt or predict how their visibility will change.
Another scenario is observer teaming. 
Observers may communicate information with each other to reach a consensus about an agent's intent.
Future work could investigate how a robot could create maximal certainty or uncertainty in a group of observers with varying levels of trustworthiness and input into the group consensus.

\subsection{Agent Modifications}
Many improvements could be made to the agent's strategy. 
Our current solution plans trajectories offline since all aspects of the environment are static. 
Real-time path planning is required to handle evolving goals, observers, and visibility regions online. 
Further investigation of observer models, particularly accurate human observer models is needed.
Our work used a simplified static observer model, but observers who are human beings or machine learning models will evolve their predictions the more examples they observe of agent behavior. 
For example, if an agent uses the same deceptive motion repeatedly, a negative observer may learn to recognize that the agent is being deceptive. 
By maintaining a dynamic model of the observer, the agent can evolve its strategy to remain unpredictable, such as by selecting unexpected decoy goals, or switching from a decoy to an ambiguous strategy.
One method for doing this is described using an RNN as the estimate of the observer model in \cite{nicholsAdversarialSamplingBasedMotion2022}.

We encourage readers to brainstorm and explore more \problemabbrev~modifications.

\section{Conclusion}
In this work, we presented and mathematically defined the \problemname~problem. 
We introduced a trajectory optimization-based method, \algorithmname, for finding trajectories that satisfy the requirements of being legible to some observers and illegible to others, while considering their visibility regions.
We demonstrate the validity of \algorithmname~on multiple environments with varying numbers of observers and types of visibility regions.
We provided multiple ways of producing illegible behavior (decoy or delay) and demonstrated the benefits and drawbacks of each method.
Our experiments show that \algorithmname~can generate trajectories that are equivalent to or outperform trajectories which maximize only for efficiency, legibility, and illegibility when observers have limited visibility or mixed motives.
Lastly, we present many interesting directions for future work on the \problemabbrev~problem.




\newpage

\bibliographystyle{splncs04}
\bibliography{dubious_planning}

\end{document}